\documentclass{article}

\usepackage{PRIMEarxiv}

\usepackage{cite}
\usepackage{amsmath,amssymb,amsfonts}
\usepackage{graphicx}
\usepackage[table,xcdraw]{xcolor}
\usepackage{colortbl}
\usepackage{textcomp}
\usepackage{multirow}
\usepackage{pdfpages}
\usepackage{subcaption}

\usepackage[utf8]{inputenc} 
\usepackage[T1]{fontenc}    
\usepackage{hyperref}       
\usepackage{url}            
\usepackage{booktabs}       
\usepackage{amsfonts}       
\usepackage{nicefrac}       
\usepackage{microtype}      
\usepackage{lipsum}
\usepackage{fancyhdr}       
\usepackage{graphicx}       
\graphicspath{{media/}}     
\usepackage{algorithm, algpseudocode}

\usepackage{amsmath}       
\usepackage{amssymb}       
\usepackage{algpseudocode} 
\usepackage{multirow}
\usepackage{booktabs}
\usepackage{bm}
\usepackage{makecell}

\usepackage{times}
\usepackage{helvet}
\usepackage{courier}
\usepackage{xcolor}

\renewcommand{\arraystretch}{1.2}

\def\BibTeX{{\rm B\kern-.05em{\sc i\kern-.025em b}\kern-.08em
    T\kern-.1667em\lower.7ex\hbox{E}\kern-.125emX}}

\algnewcommand{\Inputs}[1]{%
  \Statex \textbf{Inputs:}
  \Statex \hspace*{\algorithmicindent}\parbox[t]{.8\linewidth}{\raggedright #1}
}
\algnewcommand{\Outputs}[1]{%
  \Statex \textbf{Outputs:}
  \Statex \hspace*{\algorithmicindent}\parbox[t]{.8\linewidth}{\raggedright #1}
}
\algnewcommand{\StepOne}[1]{%
  \Statex \textbf{Step One:}
}
\algnewcommand{\StepTwo}[1]{%
  \Statex \textbf{Step Two:}
}

\title{Learning Object-Centric Representations in SAR Images with Multi-Level Feature Fusion}

\author{Ohtae Jang, Mingon Cho, Kyungtae Kim\\
    Department of Electrical Engineering, POSTECH, South Korea \\
    \texttt{$\{$otaejnag, mgcho, kkt$\}$@postech.ac.kr}
}

\begin{document}
\maketitle

\begin{abstract}
Synthetic aperture radar (SAR) images contain not only targets of interest but also complex background clutter, including terrain reflections and speckle noise. In many cases, such clutter exhibits intensity and patterns that resemble targets, leading models to extract entangled or spurious features. Such behavior undermines the ability to form clear target representations, regardless of the classifier. To address this challenge, we propose a novel object-centric learning (OCL) framework, named SlotSAR, that disentangles target representations from background clutter in SAR images without mask annotations. SlotSAR first extracts high-level semantic features from SARATR-X and low-level scattering features from the wavelet scattering network in order to obtain complementary multi-level representations for robust target characterization. We further present a multi-level slot attention module that integrates these low- and high-level features to enhance slot-wise representation distinctiveness, enabling effective OCL. Experimental results demonstrate that SlotSAR achieves state-of-the-art performance in SAR imagery by preserving structural details compared to existing OCL methods.
\end{abstract}


\section{Introduction}
\label{sec:introduction}
Synthetic Aperture Radar (SAR) imagery offers high-resolution observations under all-day, all-weather, and long-range conditions, and it has been widely applied in both military and civilian fields \cite{cumming}. As a fundamental and challenging task in remote sensing, SAR automatic target recognition (SAR-ATR) aims to automatically detect and classify objects of interest, such as vehicles, ships, or aircraft, playing a vital role in surveillance, disaster assessment, and emergency response \cite{atrnet, stamatis, zhou}. Recent advances in SAR-ATR performance have been driven by the introduction of large-scale SAR datasets \cite{atrnet, sardet100k}, along with foundation models like SARATR-X \cite{saratrx}.

SAR images for ATR typically contain not only the target but also various non-target components. Common non-target signals include background clutter caused by terrain reflections, man-made structures, or vegetation, and speckle noise stemming from sensor characteristics. In particular, due to the non-uniform scattering properties of the ground surface, background clutter can resemble target signatures in intensity and pattern \cite{clutterregion}. In such scenarios, current models tend to focus on clutter regions and learn spurious features, which hinders generalization and results in significant performance degradations under unseen background conditions \cite{irasnet, atrnet, hdanet, cfa, tsmal}. Consequently, learning to disentangle target features from background clutter has become a key challenge for improving SAR-ATR robustness and generalization.

Feature-level clutter reduction (FLCR) is one of the effective methods to address this issue \cite{irasnet}. FCLR aims to reduce background interference at the feature level while enhancing target-relevant information to refine classifier decisions. However, most existing studies rely on supervised learning, requiring precise pixel-level masks of target locations or shapes. In practice, such fine-grained annotations are rarely available in SAR imagery, and generating them is both difficult to automate and hard to ensure in terms of quality. As a result, the reliance on mask annotations limits the applicability of such methods in real-world scenarios.

To overcome these limitations, this paper proposes a novel approach to FLCR from the perspective of object-centric learning (OCL). OCL aims to decompose and learn representations of individual objects in visual scenes without manual supervision \cite{ocl1}. A key method in OCL is slot attention \cite{sa}, which enables the model to discover object-level structure in visual inputs. It introduces a set of latent representations, referred to as slots, that iteratively attend to input features via cross-attention. Through this process, each slot learns to capture the properties of a distinct object in the scene. In particular, recent advances, such as the pre-trained features \cite{dinosaur} and incorporation of slot reference frames \cite{isa}, have significantly improved the robustness and generalization of object-centric representations. These developments have laid a strong foundation for reliably learning object-centric representations and expanding the applicability of OCL to real-world data.

While OCL is inspired by human perception \cite{ocl2} and has shown promising results in optical imagery rich in color and texture cues, its application to SAR imagery presents two primary challenges due to the unique characteristics: (1) The inherent speckle noise and lack of color in SAR imagery \cite{aaaisar, cvprsar} increase the risk of slots capturing structurally irrelevant or spurious patterns, causing attention to spread over non-target regions \cite{gsa}; (2) The similarity in intensity and patterns between targets and background clutter can lead slot attention to entangle their representations.

To address these challenges, we propose SlotSAR, a novel framework that disentangles target representations from background clutter SAR images without mask annotations. We first extract high- and low-level features to obtain complementary multi-level representations. While semantic features from the pre-trained SARATR-X model \cite{saratrx} provide global context but lack local detail, local scattering features from a wavelet scattering network (WSN) \cite{wsn} highlight structurally significant regions and reduce speckle noise, yet lack semantic context, meaning that the strengths of each feature can compensate for the weaknesses of the other. We further present a multi-level slot attention (MLSA) module that integrates these high- and low-level features. The low-level features supply structural characteristics that provide fine-grained details of the targets, exhibiting distinguishable responses to targets and clutter, whereas the high-level features contribute positional information that steers the spatial allocation of attention across slots. This multi-level fusion not only guides slots toward precise target representation but also facilitates disentanglement between target and background slots.

The main contributions of this work can be summarized as follows:

\begin{itemize}\item We propose an OCL framework named SlotSAR. To the best of our knowledge, this is the first study to disentangle target and background clutter features in an unsupervised manner.

\item We present an MLSA module that integrates low-level scattering features with high-level semantic features, enhancing the preservation of structural details during slot refinement. 

\item Experimental results demonstrate that our approach achieves state-of-the-art results on both the MSTAR and ATRNet-STAR benchmarks by effectively reducing background clutter for robust target representation.

\end{itemize}

\section{Related Works}
\subsection{Target Representation in SAR Imagery.}
In SAR imagery, targets are the primary regions of interest, whereas all other areas are typically regarded as background clutter. SAR images show complicated electromagnetic scattering phenomena, including various scattering mechanisms owing to substructures on targets, clutter or interfering signals, and speckle noise \cite{jongsenlee, Frost}. Given these complexities, it is essential to extract target representations that can effectively disentangle target structures from complex scattering patterns. A target representation denotes an abstracted form of information that encapsulates the structural, scattering, spatial, and morphological characteristics of the target \cite{potter1997attributed, zhang2023vsfa, zhao2024adaptive, busip, ifts}. Therefore, by effectively separating targets from background clutter, the model can rely on target information for decision-making, improving generalization performance \cite{irasnet}.

Early approaches primarily relied on statistical thresholding based on pixel brightness to capture target regions, which were then used as inputs for feature extraction \cite{ifts, lmbncnn, tsmal}. Subsequent studies improved model robustness against clutter by selectively augmenting clutter regions and applying contrastive learning \cite{cfa}. More recent methods have focused on enhancing the signal-to-clutter ratio (SCR) in the feature space and reducing background clutter during learning to better capture target representations. \cite{irasnet, hdanet}. However, most existing methods rely heavily on ground-truth masks to identify target regions, which limits their scalability and practicality in real-world applications where such annotations are often unavailable. In contrast, our method is designed to operate effectively under complex background conditions while capturing fine-grained structural characteristics of targets in SAR imagery without relying on ground-truth masks.

\subsection{Object-Centric Learning.}
OCL aims to decompose a scene into independent representations of individual objects \cite{dinosaur, smsa}. A representative method in this field is slot attention \cite{sa}, which clusters features into object-centric representations by performing cross-attention between input features and a set of slots. Each slot learns to encode the properties of a distinct object in the scene. OCL can significantly enhance the generalization of vision models by aligning with the causal structure of the physical world \cite{scholkopf2021toward,dittadi2021generalization}. Subsequent studies have extended slot attention in various directions, including incorporation of slot reference frames \cite{isa}, novel encoder or decoder architectures \cite{dinosaur, jiang2023object, wu2023slotdiffusion}, multi-modal learning \cite{kim2023improving, jsa}, and application across domains such as medical image \cite{liao2025future} and video modality \cite{kipf2021conditional, gsa}. In the field of remote sensing, slot attention has been adapted to a range of applications such as SAR–hyperspectral fusion \cite{yu2024multimodal} and audio-image multimodal learning \cite{han2025slotfusion}. 

Notably, Kim et al. introduced a top-down pathway that maps each slot to the nearest code in a finite codebook, providing semantic guidance for improved object decomposition in complex real-world scenes \cite{smsa}. However, in SAR imagery, the statistical similarity between targets and background clutter \cite{atrnet} often leads to subtle inter-slot differences, resulting in overlap in assigned codes during vector quantization. To address this limitation, we introduce a multi-level fusion module that integrates structural details and semantic context, facilitating disentanglement between target and background slots.

\begin{figure*}[t]
\centering
\includegraphics[width=0.9\linewidth]{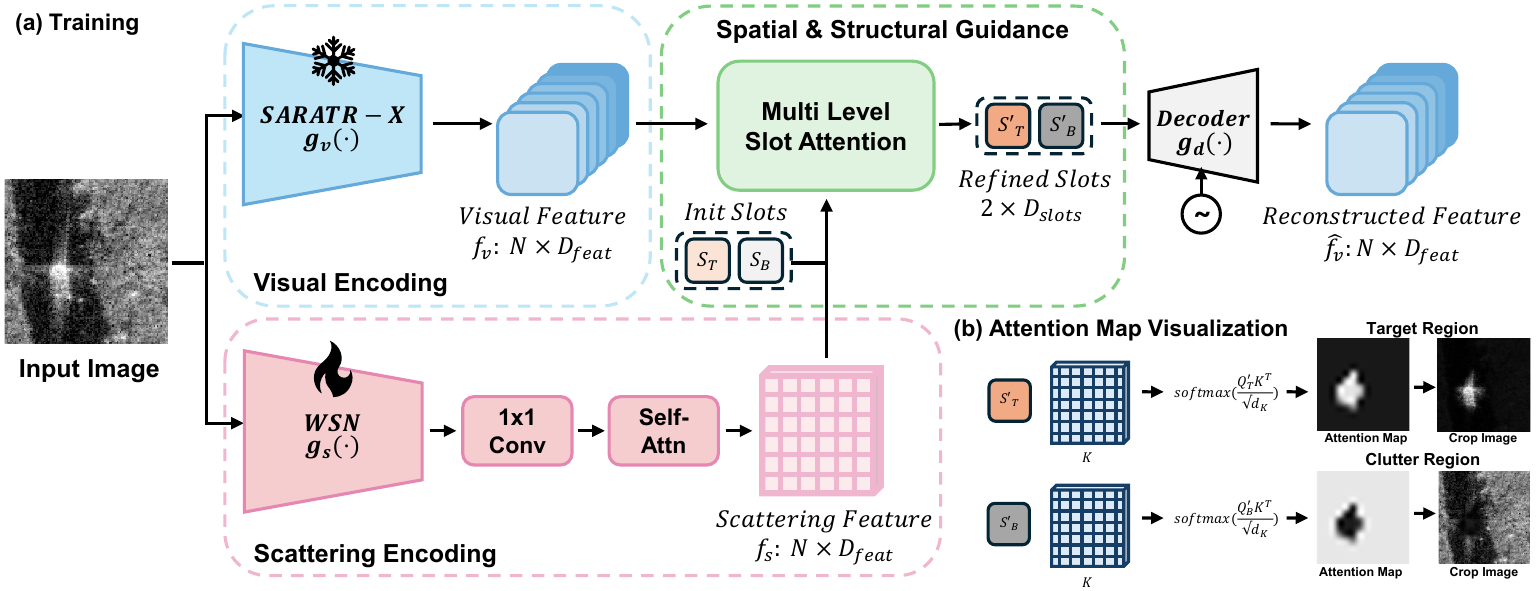}
\caption{Overall pipeline of the SlotSAR framework. Two types of features are extracted from the input image and fed into the MLSA module, which disentangles target and background clutter slots by complementing each other.}
\label{fig:SlotSAR}
\vspace{-1.3em}
\end{figure*}

\section{Methodology}
The core objective of SlotSAR is to achieve clear disentanglement between target and background clutter slots, enabling robust OCL under the complex characteristics of SAR imagery. By effectively separating these two representations, SlotSAR allows downstream models to focus solely on target information, which is known to improve generalization in ATR tasks \cite{irasnet}. Fig.~\ref{fig:SlotSAR} illustrates the overall pipeline of the proposed SlotSAR framework. The visual features are extracted using SARATR-X. In parallel, low-level scattering features are obtained using a WSN, which captures the intrinsic scattering characteristics of the target, exhibiting distinguishable responses to targets and clutter \cite{wsn, busip}. The extracted features are then fed into an MLSA module, where slot representations are iteratively updated through attention. In the first stage, the visual features are used to estimate coarse attention maps that approximate the spatial location of the target. Next, the scattering features are combined with the coarse attention maps to refine the slots and improve disentanglement between target and background clutter. The final refined slots are then decoded to reconstruct the input image and generate slot-wise masks.

\subsection{Features Encoding}
\subsubsection{Scattering Encoding.}
To accurately capture the local scattering characteristics of SAR imagery, SlotSAR incorporates a WSN based on the wavelet transform (WT) \cite{WT}. WT is a major mathematical tool for image analysis due to its multi-scale and localization properties. These strengths make WT particularly effective for analyzing non-smooth signals and complex textures, which are commonly observed in SAR images.

SlotSAR adopts a 2D Morlet wavelet to extract elliptical and circular scattering patterns frequently found in SAR scenes \cite{wsn}.
To further increase modeling flexibility, WSN replaces fixed wavelet filters with a parameterized filter bank of Morlet wavelets. The generalized form of the learnable Morlet wavelet is:
\begin{align}
    \psi_{\sigma, \theta, \xi, \gamma}(u) ={}& 
    e^{\left(-\frac{ \| D_\gamma R_\theta(u) \|^2 }{2\sigma^2} \right)} \left( e^{(i \xi u') - \beta} \right),
\label{eq:morlet}
\end{align} where $R_\theta$ is the rotation matrix with angle $\theta$, and $D_\gamma$ is the anisotropic dilation matrix controlling the aspect ratio. The term $u' = u_1 \cos \theta + u_2 \sin \theta$ aligns the wavelet to the specified orientation, and the constant $\beta$ ensures a zero-mean condition. The wavelet parameters, scale ($\sigma$), orientation ($\theta$), frequency ($\xi$), and aspect ratio ($\gamma$), are learned during training. This allows the model to adaptively generate wavelet bases tailored to SAR-specific scattering patterns. As a result, the WSN effectively captures low-level scattering structures \cite{wsn}, suppresses speckle noise while preserving structural features \cite{Duan, aaaisar}, and facilitates discrimination between target regions and background clutter based on their statistical differences \cite{Tello, Grandi1, Grandi2}.

For this reason, scattering features are well-suited for OCL, despite the speckle noise and lack of color information in SAR images. Let $I \in \mathbb{R}^{H \times W}$ denote the input SAR image of height $H$ and width $W$. The scattering features are extracted via the WSN $g_s$, resulting in: \begin{align}
     F_s = g_s(I), \quad F_s \in \mathbb{R}^{C \times H' \times W'}, 
\label{eq:scattering_encoder}
\end{align}with $C$, $H'$, $W'$ as channel, height, and width. To enable token-wise modeling, the feature map $F_s$ is reshaped into $\mathbb{R}^{N \times D_s}$ by pooling, resize, and flattening, where $N$ and $D_s$ denote the number of tokens and the feature dimension, respectively. These scattering features are subsequently refined through a lightweight encoder to capture the correlations between different scattering channels, followed by a self-attention module:
\begin{equation}
\begin{aligned}
    f_e(F) &= \text{ReLU}(\text{BN}(f^{1 \times 1}(F))), \\ \quad F_s' &= f_e(f_e(\text{BN}(F_s))), \\ \quad F_s'' &= \text{SelfAttention}(F_s'), 
\end{aligned}
\label{eq:scattering_encoder2}
\end{equation} where BN denotes batch normalization, and $f^{1 \times 1}$ represents a convolution operation with a filter size of $1 \times 1$. As a result, the final scattering feature $F_s'' \in \mathbb{R}^{N \times D_s}$ highlights structural features and encodes discriminative cues for distinguishing targets from clutter, providing a robust complement to the visual features in SlotSAR. In the MLSA module, $F_s''$, derived from scattering features, offers structural guidance that helps slots focus on target-relevant regions.

\subsubsection{Visual Encoding.}
Given an input SAR image $I \in \mathbb{R}^{H \times W}$, we employ the SARATR-X foundation model $g_v$ to extract semantic features as: \begin{align}
F_v = g_v(I), \quad F_v \in \mathbb{R}^{N \times D_{\text{feat}}}, 
\label{eq:visencoder}
\end{align}where $D_{\text{feat}}$ denotes the feature dimensionality.
SARATR-X is pretrained via self-supervised masked patch reconstruction, enabling the model to learn high-level semantic features from SAR imagery. Multi-scale gradient features are then used to refine these representations by suppressing speckle noise and emphasizing structural cues such as edges and target contours. The high-level semantic features are fed into the MLSA module to provide coarse spatial guidance for OCL.

\subsection{Multi-Level Slot Attention}
\subsubsection{Slot Initialization and Feature Projection.} 
We first designate one of the two slots $S$ as the target slot $S_T$ and the other as the background clutter slot $S_B$, such that $S = [S_T, S_B] \in \mathbb{R}^{2 \times D_{\text{slot}}}$ with $S_T, S_B \in \mathbb{R}^{D_{\text{slot}}}$, where $D_{\text{slot}}$ denotes the dimensionality of slots. We first define the initial slots $S_{init} \sim \mathcal{N}(\mu, \text{diag}(\sigma))$ sampled from a Gaussian distribution, where $\mu, \sigma \in \mathbb{R}^{D_{\text{slot}}}$ are learnable parameters.

After initializing the slots, we transform the input features through normalization and projection as follows:\begin{equation}
\begin{aligned}
F_s'' &= \text{MLP}(\text{LayerNorm}(F_s'')) \in \mathbb{R}^{N \times D_s}, \\
F_v &= \text{MLP}(\text{LayerNorm}(F_v)) \in \mathbb{R}^{N \times D_{\text{feat}}}.
\end{aligned}
\label{eq:encoder2}
\end{equation} To perform attention between the slots and the visual features, the key and value representations are obtained using the following equations: \begin{equation}
\begin{aligned}
K &= W^k \cdot \text{LayerNorm}(F_v) \in \mathbb{R}^{N \times D_{\text{slot}}}, \\
V &= W^v \cdot \text{LayerNorm}(F_v) \in \mathbb{R}^{N \times D_{\text{slot}}},
\end{aligned}
\label{eq:keyvalue}
\end{equation}where $W^k$ and $W^v$ are the linear projection layers that map the features to the $D_{\text{slot}}$-dimensional embedding space.

The slots are iteratively refined at each step $t = 1, \dots, T$, allowing them to capture representations corresponding to objects from the input features $F_v$. To update the slots, a query $Q_t$ is computed at each iteration $n$ by projecting the layer-normalized slots from the previous step: 
\begin{align}
Q_t = W^q \cdot \text{LayerNorm}(S_{t-1}) \in \mathbb{R}^{2 \times D_{\text{slot}}}, 
\label{eq:query}
\end{align} where $W^q$ are the linear projection layers.

\subsubsection{Top-Down Spatial Information.} 
Similar to vision images, we observed that using the $g_v$ encoder for SAR imagery also enables the slot attention module to produce slots that encode rough semantic information about objects. Inspired by the self-modulating slot attention \cite{smsa} that leverages top-down information, we utilize this coarse object-level information to extract top-down spatial cues from the image. 

The top-down spatial cues are obtained via the slot attention map $A \in \mathbb{R}^{2 \times N}$, which is computed through the dot-product between the query and key representations:
\begin{align}
\quad A_{i,j} = \frac{\exp(P_{i,j})}{\sum_{l=1}^2 \exp(P_{l,j})},
\label{eq:attnmap}
\end{align}
where $P = K (Q_t)^\top / \sqrt{d_h}$, $i$ is the key index, $j$ is the slot index, and $\sqrt{d_h}$ is a scaling factor to normalize the dot-product. These slot attention map $A$ reflect where objects are located within the image. To emphasize regions that are more likely to contain objects, the attention map is shifted to have a mean value of 1 to construct the spatial map: \begin{align}
M_s = 1 + (A - \bar{A}) \in \mathbb{R}^{2 \times N},
\label{eq:shifting}
\end{align} where $\bar{A}$ denotes the average of the attention map.

\subsubsection{Fusion Map.}
However, SAR images exhibit high similarity between targets and background clutter in terms of intensity and texture. When existing methods are employed \cite{smsa}, this similarity leads target and background slots to collapse into the same code, making it difficult to extract reliable top-down semantic information. Therefore, we introduce a simple yet effective fusion with scattering features. 

This fusion aims to inject structural detail into the slot representation process, effectively combining coarse semantic cues with fine-grained structural information to better disentangle target and clutter regions \cite{unet}. As low-level features introduced during encoding level fusion can disrupt semantic abstraction and cause unreliable slot operation, we perform the fusion within the slot attention module to maintain high-level representation \cite{dinosaur}. The fusion is implemented through the following fusion map: \begin{align}
M_f = F_s'' \otimes M_s \in \mathbb{R}^{2 \times N \times D_s}.\label{eq:guidance}
\end{align} The fusion map $M_f$ is computed as the outer product of the spatial map and the scattering feature. This operation reflects the idea that the resulting scores simultaneously capture both spatial and structural details in the approximate regions where the target is likely to be located. Subsequently, the fusion map $M_f$ refines $V$ through element-wise multiplication, emphasizing target-relevant features with detailed spatial structure. Afterwards, we perform a dot-product with the normalized attention map $\hat{A}_{i,j} = A_{i,j}/\sum_{l=1}^N A_{i,l}$. This process reduces clutter focus while enhancing the target response, allowing the representation to incorporate fine-grained details of the target.

The updated slot $\bar{S}_t$ is then computed using a gated recurrent unit (GRU) \cite{gru}, where $S_{t-1}$ is used as the hidden state and the input is given by $\hat{A}^\top \cdot (M_f \odot V)$: \begin{align}
\bar{S}_t = \text{GRU}(S_{t-1}, \hat{A}^\top \cdot (M_f \odot V)).
\label{eq:zs}
\end{align} Finally, the slots are updated via MLP:
\begin{align}
S_t = \text{MLP}(\text{LayerNorm}(\bar{S}_t)) \in \mathbb{R}^{2 \times N}.
\label{eq:slotmlp}
\end{align} 
After $T$ iterations, the target and background clutter representations are disentangled, resulting in the refined slots $S_T'$ and $S_B'$. 

\subsubsection{Training}
The refined target and background slots $S_T'$, $S_B'$ are passed through a lightweight MLP decoder $g_d(\cdot)$. The reconstruction loss encourages the decoded output $\hat{f_v}$ to match the original visual features $f_v$ extracted from the encoder:
\begin{align}
\mathcal{L}_{\text{Recon}} = | f_v - g_d(S_T'; S_B') |_2,
\label{eq:lrecon}
\end{align} which has been shown to provide more robust training signals for real-world datasets \cite{dinosaur}.

\begin{table*}[t]
\centering
\renewcommand{\arraystretch}{1.5} 
\resizebox{0.99\linewidth}{!}{
\begin{tabular}{c|ccccccc|ccccccc}
\toprule
\toprule
\multirow{3}{*}{Setting} & \multirow{2}{*}{\makecell{Slot\\Attention}} & \multicolumn{2}{c}{DINOSAUR} & \multicolumn{2}{c}{ISA}    & SMSA    & \multirow{2}{*}{\makecell{SlotSAR\\(Ours)}} & \multirow{2}{*}{\makecell{Slot\\Attention}} & \multicolumn{2}{c}{DINOSAUR} & \multicolumn{2}{c}{ISA}    & SMSA    & \multirow{2}{*}{\makecell{SlotSAR\\(Ours)}} \\ \cmidrule(lr){3-7} \cmidrule(lr){10-14}
                          &                             & DINOv2         & SARATRX       & DINOv2    & SARATRX          & SARATRX &                          &                             & DINOv2         & SARATRX       & DINOv2    & SARATRX          & SARATRX &                          \\ \cmidrule(lr){2-15} 
                          & \multicolumn{7}{c|}{Adjusted Random Index (ARI)}                                                                             & \multicolumn{7}{c}{Mean Best Overlap (mBO)}                                                                                  \\ \midrule
SOC-40                    & 0.79\%                      & 0.50\%       & 23.00\%       & 11.82\% & 36.86\%          & 0.50\%  & \textbf{43.65\%}         & 8.14\%                      & 6.07\%       & 22.77\%       & 16.51\% & 32.21\%          & 6.07\%  & \textbf{36.81\%}         \\ \midrule
SOC-50                    & 16.58\%                     & 3.43\%       & 25.04\%       & 11.79\% & 26.82\%          & 3.43\%  & \textbf{37.57\%}         & 15.95\%                     & 5.41\%       & 23.47\%       & 15.44\% & 24.50\%          & 5.41\%  & \textbf{32.10\%}         \\ \midrule
EOC-Azi.                    & 4.46\%                      & 0.39\%       & 25.36\%       & 11.14\% & 34.87\%          & 0.39\%  & \textbf{43.07\%}         & 10.04\%                     & 6.05\%       & 24.51\%       & 16.25\% & 30.52\%          & 6.05\%  & \textbf{37.07\%}         \\
- (60)                     & 3.85\%                      & 0.51\%       & 22.24\%       & 9.45\%  & 30.97\%          & 0.51\%  & \textbf{41.11\%}         & 9.42\%                      & 5.61\%       & 22.09\%       & 14.70\% & 27.13\%          & 5.61\%  & \textbf{35.19\%}         \\
- (120)                    & 4.98\%                      & 0.02\%       & 29.09\%       & 13.37\% & 36.65\%          & 0.02\%  & \textbf{44.19\%}         & 11.18\%                     & 7.02\%       & 27.99\%       & 18.77\% & 32.68\%          & 7.02\%  & \textbf{38.81\%}         \\
- (180)                    & 4.41\%                      & 0.27\%       & 24.35\%       & 9.98\%  & 36.72\%            & 0.27\%  & \textbf{43.56\%}       & 9.55\%                      & 5.33\%       & 22.97\%       & 14.67\% & 31.38\%          & 5.33\%  & \textbf{36.84\%}                  \\
- (240)                    & 3.98\%                      & 1.03\%       & 22.61\%       & 9.84\%  & 33.25\%          & 1.03\%  & \textbf{42.92\%}         & 9.15\%                      & 5.46\%       & 22.07\%       & 14.66\% & 28.43\%          & 5.46\%  & \textbf{36.09\%}         \\
- (300)                    & 5.57\%                      & 0.05\%       & 28.79\%       & 13.23\% & 37.06\%          & 0.05\%  & \textbf{43.73\%}         & 11.57\%                     & 7.28\%       & 28.14\%       & 18.99\% & 33.27\%          & 7.28\%  & \textbf{38.50\%}         \\ \midrule
EOC-Band                  & 22.91\%                     & 0.35\%       & 22.95\%       & 15.60\% & 36.51\%          & 0.35\%  & \textbf{45.27\%}         & 21.88\%                     & 5.44\%       & 22.25\%       & 17.71\% & 31.44\%          & 5.44\%  & \textbf{37.58\%}         \\ \midrule
EOC-Dep.                    & 12.32\%                     & 0.40\%       & 21.72\%       & 12.79\% & 38.81\%          & 0.40\%  & \textbf{41.67\%}         & 13.91\%                     & 5.98\%       & 22.04\%       & 17.08\% & 33.62\%          & 5.98\%  & \textbf{35.21\%}         \\
- (30)                     & 12.59\%                     & 0.37\%       & 18.33\%       & 10.60\% & 38.70\%          & 0.37\%  & \textbf{40.91\%}         & 13.79\%                     & 5.65\%       & 19.56\%       & 15.24\% & 33.25\%          & 5.65\%  & \textbf{34.43\%}         \\
- (45)                     & 12.21\%                     & 0.48\%       & 21.86\%       & 12.32\% & 39.91\%          & 0.48\%  & \textbf{42.80\%}         & 13.88\%                     & 6.02\%       & 22.04\%       & 16.81\% & 34.50\%          & 6.02\%  & \textbf{36.17\%}         \\
- (60)                     & 11.71\%                     & 0.33\%       & 25.07\%       & 15.25\% & 37.82\%          & 0.33\%  & \textbf{41.29\%}         & 13.63\%                     & 6.10\%       & 24.40\%       & 18.78\% & 33.09\%          & 6.10\%  & \textbf{35.03\%}         \\ \midrule
EOC-Pol.                    & 17.31\%                     & 0.40\%       & 28.91\%       & 15.04\% & 36.26\%          & 0.40\%  & \textbf{43.69\%}         & 17.15\%                     & 6.08\%       & 26.82\%       & 18.43\% & 31.67\%          & 6.08\%  & \textbf{37.01\%}         \\
- (VV)                     & 17.29\%                     & 0.58\%       & 31.68\%       & 15.28\% & 36.06\%          & 0.58\%  & \textbf{44.44\%}         & 16.98\%                     & 5.87\%       & 28.49\%       & 18.20\% & 31.32\%          & 5.87\%  & \textbf{37.54\%}         \\
- (HV)                     & 17.31\%                     & 0.38\%       & 27.48\%       & 14.84\% & 36.42\%          & 0.38\%  & \textbf{43.37\%}         & 17.08\%                     & 6.04\%       & 25.73\%       & 18.20\% & 31.88\%          & 6.04\%  & \textbf{36.76\%}         \\
- (VH)                     & 17.21\%                     & 0.20\%       & 27.66\%       & 14.95\% & 36.29\%          & 0.20\%  & \textbf{43.28\%}         & 17.21\%                     & 6.19\%       & 26.09\%       & 18.56\% & 31.80\%          & 6.19\%  & \textbf{36.71\%}         \\ \midrule
EOC-Scene                 & 1.29\%                      & 0.28\%       & 17.24\%       & 12.83\% & 25.65\%          & 0.28\%  & \textbf{30.85\%}         & 8.92\%                      & 6.33\%       & 19.89\%       & 17.51\% & 24.69\%          & 6.33\%  & \textbf{28.47\%}         \\
- (City)                   & 0.57\%                      & 0.33\%       & 14.16\%       & 8.72\%  & 23.71\%          & 0.33\%  & \textbf{26.54\%}         & 9.35\%                      & 7.03\%       & 18.94\%       & 16.16\% & 24.15\%          & 7.03\%  & \textbf{26.03\%}         \\
- (Factory)                & 1.88\%                      & 0.22\%       & 20.48\%       & 16.43\% & 27.06\%          & 0.22\%  & \textbf{35.40\%}         & 8.39\%                      & 5.72\%       & 20.96\%       & 18.61\% & 24.83\%          & 5.72\%  & \textbf{31.02\%}         \\
- (Woodland)               & 4.87\%                      & 1.15\%       & 29.51\%       & 28.85\% & 34.68\%          & 1.15\%  & \textbf{39.85\%}         & 9.23\%                      & 4.28\%       & 25.65\%       & 24.84\% & 29.34\%          & 4.28\%  & \textbf{33.66\%}         \\ \midrule
AVG.                      & 9.24\%                      & 0.56\%       & 24.17\%       & 13.53\% & 34.34\%          & 0.56\%  & \textbf{40.91\%}         & 12.69\%                     & 5.95\%       & 23.66\%       & 17.43\% & 30.27\%          & 5.95\%  & \textbf{35.10\%}         \\ \bottomrule
\end{tabular}%
}
\caption{Quantitative comparison of the proposed SlotSAR framework with seven existing methods across 21 evaluation settings under standard operating conditions (SOC) and extended operating conditions (EOC) on the ATRNet-STAR dataset.}
\label{tab:comparison}
\end{table*}

\section{Experimental Results}
\subsection{Experimental Setup}
\textbf{Datasets.} 
To evaluate the effectiveness of the proposed model, we performed extensive experiments on the large-scale benchmark ATRNet-STAR dataset \cite{atrnet}. The dataset comprises 40 vehicle-based target classes and 194,324 annotated SAR images, collected under complex imaging conditions, including Ku/X bands, various polarizations, diverse scenes, and variations in azimuth angle, depression angle, and target position, which define the key factors of extended operating conditions (EOC). In particular, the dataset provides bounding box annotations for targets, which we used in combination with \cite{ifts} to develop a labeling tool to generate binary segmentation maps in pixels, allowing quantitative evaluation of OCL results. Further details are provided in the supplementary material.

\textbf{Evalutation Metric.} We evaluate the task using three metrics: adjusted rand index (ARI), mean best overlap (mBO), and mean intersection over union (mIoU). ARI measures clustering similarity; mBO reflects the maximum overlap with ground truth target regions (excluding background); and mIoU quantifies the average overlap between targets and background, enabling us to assess the model’s ability to disentangle targets from clutter.
In quantitative evaluation, following previous work \cite{sa, dinosaur}, we evaluated our method by comparing the alpha masks generated by the decoder for each slot (target and background clutter) with the ground truth binary segmentation maps.
In the ablation study, to evaluate whether each slot effectively captures its corresponding target region, we compare binary masks derived from attention maps with the ground-truth segmentation maps. For the mBO and mIoU computations, we apply the Hungarian matching algorithm to align the predicted masks with the ground truth.

\begin{figure*}[t]
\centering
\includegraphics[width=0.95\linewidth]{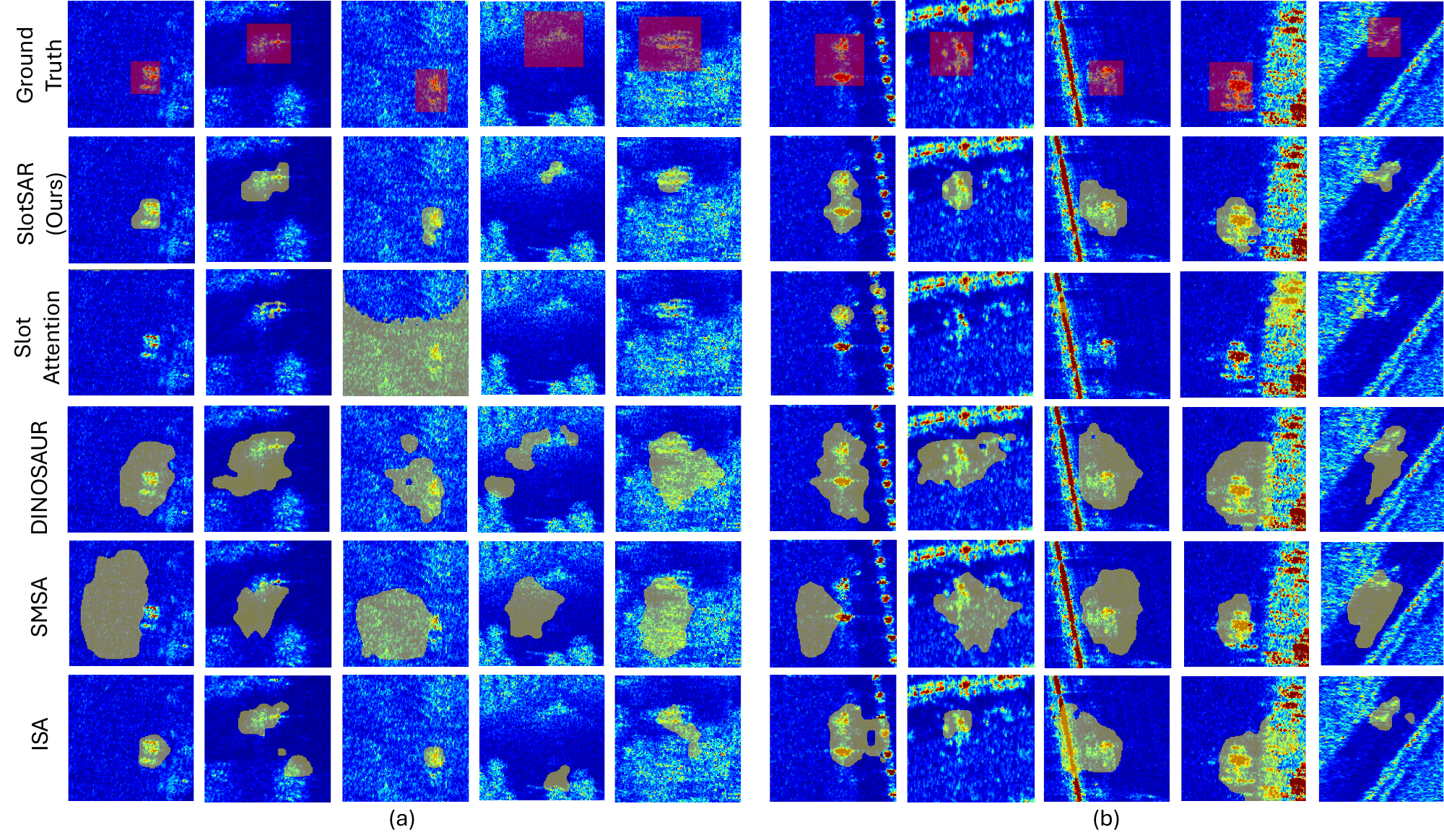}
\caption{Example results on the ATRNet-STAR dataset. (a) shows results under natural clutter conditions, while (b) presents results in scenes with significant clutter caused by man-made structures. The red regions indicate the ground-truth target regions, while the yellow target regions represent the areas predicted by the model.}
\label{fig:comparison}
\end{figure*}

\textbf{Implementation Details.}
SlotSAR is trained using the Adam optimizer \cite{adam} with a learning rate of 7e-4, linearly warmed up over 10k steps and then exponentially decayed, with gradient clipping (threshold 1) applied to stabilize training. The model is trained for 200 epochs with a batch size of 64. Additionally, the decoder is implemented as an MLP. We set the slot dimension to $D_{\text{slot}} = 256$, perform $T = 3$ iterative refinements in MLSA, and use feature token representations with $D_s = 64$, $N = 196$, and $D_{\text{feat}} = 512$. All experiments are conducted on a single NVIDIA RTX 4090 GPU (24 GB) with an Intel Xeon Gold 6240 CPU and 512 GB RAM.

Our model is trained in two stages, following the methodology used in prior studies \cite{gsa}. First, to prevent overfitting and stabilize the initial representation learning, we pretrain the model on the training set of the EOC-scene dataset, which is a saliency dataset characterized by relatively simple background clutter. This helps the model effectively learn object-centric features. Then, we fine-tune the entire model using the training set of each scenario-specific dataset. This fine-tuning phase enhances generalization and stabilizes convergence under more complex settings, and is applied consistently across all models.

\subsection{Quantitative Results}
We compare the performance of the proposed SlotSAR framework with several state-of-the-art OCL methods: Slot Attention \cite{sa}, DINOSAUR \cite{dinosaur}, Invariant Slot Attention (ISA) \cite{isa}, and Self-Modulating Slot Attention (SMSA) \cite{smsa}. For fair comparison, we evaluate each method using both DINOv2 \cite{oquab2023dinov2} and SARATR-X, with a shared decoder architecture across all models. As shown in Tab.~\ref{tab:comparison}, our method achieves state-of-the-art performance across challenging operating conditions. Specifically, SlotSAR improves the average ARI and mBO by +6.57$\%$ and +5.83$\%$, respectively, compared to the best-performing baselines. The EOC-scene setting poses a significant challenge, as the model is trained on simple background clutter but evaluated in much more complex environments. These include woodland scenes with severe speckle noise from natural clutter, and industrial areas such as factories, where strong reflections from man-made structures introduce artificial clutter. Despite these difficult conditions, SlotSAR demonstrates robust performance, achieving ARI improvements of +10.34$\%$ and +14.92$\%$ over DINOSAUR, and +5.17$\%$ and +8.4$\%$ over ISA in the woodland and factory scenes, respectively. These results indicate that the structural guidance provided by the scattering features effectively enhances target–clutter disentanglement in SAR imagery, even in previously unseen complex environments.

\subsection{Qualitative Results}
Fig.~\ref{fig:comparison} presents a qualitative comparison between the baseline and the proposed models under various scenarios. The main objective of our method is to assess how effectively the model can disentangle target features. To ensure fair comparison, we visualize attention maps as binary masks for all models, allowing us to observe how well each slot focuses on the target. The evaluation includes a wide range of cases, ranging from scenes with minimal to severe natural clutter (Fig.~\ref{fig:comparison}(a)) and from low to high levels of artificial clutter such as man-made structures (Fig.~\ref{fig:comparison}(b)).
From the results, we can observe the following: 1) when interference from natural clutter is dominant, the proposed MLSA module enables more robust focus on both the location and fine structural details of the target; and 2) in scenarios where artificial structures exhibit stronger backscatter than the target, the proposed SlotSAR framework more reliably disentangles the target from the background clutter. These results demonstrate that the proposed model is capable of effectively disentangling targets from background clutter in SAR imagery, even under challenging conditions.

\begin{figure}[hbt!]
\centering
\includegraphics[width=0.65\linewidth]{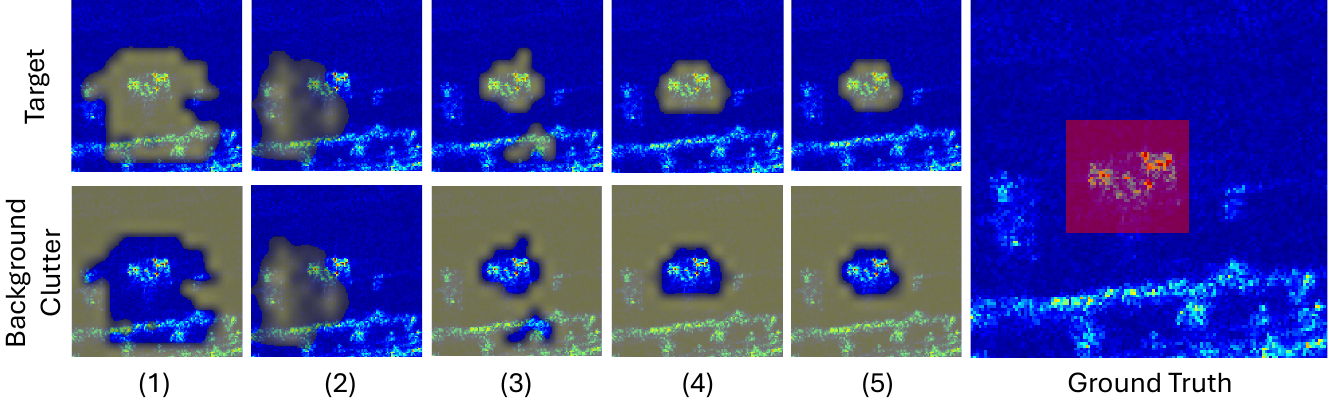}
\caption{Slot attention map visualizations of target and background clutter according to the index settings in Tab.~\ref{tab:ablation_component}.}
\label{fig:ablation_component}
\end{figure}

\subsection{Ablation Study}
\begin{figure}[hbt!]
\centering
\begin{minipage}[hbt!]{0.6\linewidth}   
\textbf{SlotSAR Components.} Tab.~\ref{tab:ablation_component} presents the ablation results for the scattering feature ($F_s$) and the spatial map ($M_s$), and Fig.~\ref{fig:ablation_component} offers the corresponding visual comparisons. We also include an analysis of the effect of VQ in SAR imagery. The first row shows the DINOSAUR without any modules. The second row corresponds to applying VQ and $M_s$, equivalent to the SMSA structure. The last row includes both $F_s$ and Ms, representing the SlotSAR framework. From the second row and Fig.~\ref{fig:ablation_component}, we observe that applying VQ leads to slots being mapped to a common codebook, resulting in diminished competition among slots and focus on identical spatial regions. The third row uses spatial guidance via $M_s$ to indicate approximate object locations. The fourth row evaluates the effect of incorporating the $F_s$ alone, and the results indicate that the structural guidance provides substantial improvements in both quantitative metrics and visual separation between target and clutter. Finally, combining $M_s$ and $F_s$ results in more precise localization and better separation between target and clutter, highlighting the complementary roles of $M_s$ and $F_s$ in SlotSAR.
\end{minipage}%
\hfill
\begin{minipage}[hbt!]{0.37\linewidth}   
\centering
\renewcommand{\arraystretch}{1.3} 
\resizebox{\linewidth}{!}{
\begin{tabular}{cccc|ccc}
\toprule
\toprule
\multirow{2}{*}{Index} & \multicolumn{3}{c|}{Component} & \multicolumn{3}{c}{Metric} \\ \cline{2-7} 
 & VQ & $\bm{M_s}$ & $\bm{F_s}$ & ARI & mBO & mIOU \\ \midrule
(1) & \multicolumn{3}{c|}{DINOSAUR (Baseline)} & 16.20\% & 18.86\% & 47.17\% \\
(2) & \checkmark & & \checkmark & 0.28\% & 6.33\% & 46.72\% \\ \midrule
(3) & & \checkmark & & 29.34\% & 26.58\% & 56.08\% \\
(4) & & & \checkmark & 31.16\% & 27.46\% & 57.83\% \\ \midrule
(5) & & \checkmark & \checkmark & \textbf{33.36\%} & \textbf{29.19\%} & \textbf{58.46\%} \\
\bottomrule
\end{tabular}}
\captionof{table}{OCL performance of the SlotSAR framework under various configurations involving 
the spatial map ($M_s$) and the scattering feature ($F_s$). Vector quantization (VQ), although not part of 
SlotSAR, is additionally evaluated as a separate comparative baseline.}
\label{tab:ablation_component}
\end{minipage}
\end{figure}

\begin{figure}[hbt!]
\centering
\begin{minipage}[hbt!]{0.6\linewidth}   
\textbf{Impact of Iterative Refinement in Slot Attention.} We set the number of iterations to $T$=3, where the guidance $M_f$ is most effectively utilized, and analyze how the guidance signal is progressively refined at each iteration. As shown in Fig.~\ref{fig:ablation_iter}, the baseline model tends to maintain its initial prediction even after multiple iterations, failing to recover fine structural details. We also observe that the attention sometimes expands into clutter regions that have similar or even stronger backscatter intensity than the actual target. In contrast, the proposed model begins to localize the approximate target region as early as $T$=1 and gradually refines the segmentation by leveraging both the scattering features
\end{minipage}%
\hfill
\begin{minipage}[hbt!]{0.37\linewidth}   
\centering
\renewcommand{\arraystretch}{1.4} 
\resizebox{0.95\linewidth}{!}{
\begin{tabular}{c|c|cccc}
\toprule
\toprule
\multirow{2}{*}{Testset}   & \multirow{2}{*}{$T$} & \multicolumn{2}{c}{ARI}     & \multicolumn{2}{c}{mBO}     \\ \cline{3-6} 
                                    &                             &   Baseline   &   SlotSAR   &   Baseline   &   SlotSAR   \\ \midrule
\multirow{3}{*}{EOC-Scene} & \textbf{1}                  & 11.69\%           & \textbf{20.76\%} & 16.23\%           & \textbf{21.74\%} \\
                                    & \textbf{2}                  & 12.53\%           & \textbf{32.23\%} & 16.82\%           & \textbf{28.63\%} \\
                                    & \textbf{3}                  & 16.20\%           & \textbf{33.36\%} & 18.86\%           & \textbf{29.19\%} \\ \bottomrule
\end{tabular}%
}
\captionof{table}{Performance comparison between DINOSAUR (baseline) and SlotSAR across slot iterative refinement.}
\label{tab:ablation_iter}
\end{minipage}
\end{figure} \vspace{-1em} and spatial priors. As shown in Tab.~\ref{tab:ablation_iter}, our method shows a 12.6$\%$ improvement in ARI over iterations. These results indicate that the superior performance of our method is primarily attributed to the effectiveness of the structural and spatial guidance provided in the refinement process.

\begin{figure}[t]
\centering
\includegraphics[width=0.55\linewidth]{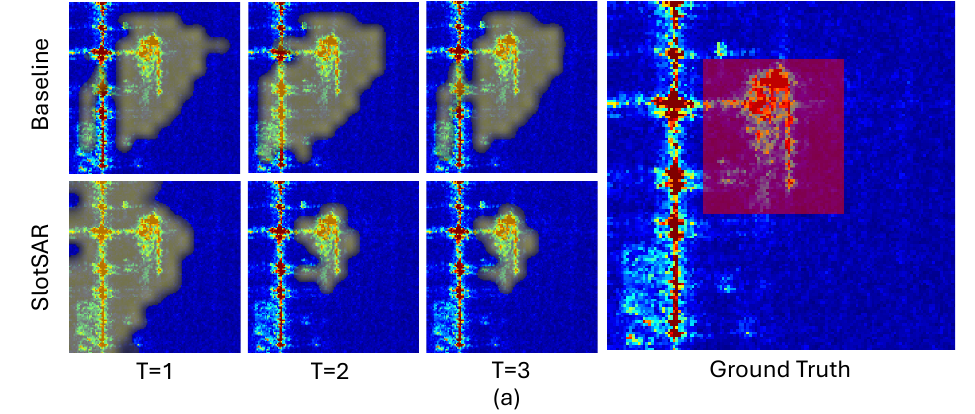}
\caption{Slot attention map visualizations of DINOSAUR (baseline) and SlotSAR across slot iterative refinement.}
\label{fig:ablation_iter}
\end{figure}

\begin{figure}[hbt!]
\centering
\includegraphics[width=0.6\linewidth]{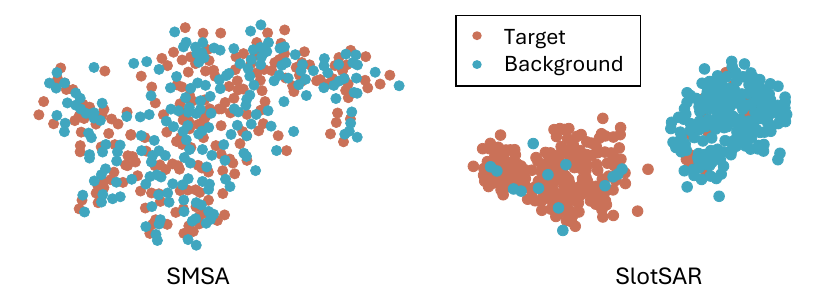}
\caption{t-SNE visualizations \cite{tsne} of target and background clutter slots for SMSA and SlotSAR.}
\label{fig:tsne}
\end{figure}

\textbf{Visualizing Slot Representations in SAR Imagery.}
To verify how effectively the target and background clutter are separated in the slot, we visualized the features of each slot using t-SNE \cite{tsne} and compared them with the modulating-based method. As shown in Fig.~\ref{fig:tsne}, the slot-wise distributions exhibit significant overlap in SMSA \cite{smsa}, whereas the proposed SlotSAR achieves clear slot-level separation, even with modulation. This suggests that the introduction of WSN effectively enables distinction in situations where semantic feature learning based on a codebook is challenging due to the visual similarity between targets and backgrounds in SAR images.

\section{Conclusion}
In this study, we presented a novel SlotSAR framework, which represents the first unsupervised target and clutter disentanglement approach in SAR imagery. Unlike conventional approaches that rely on mask supervision, SlotSAR introduces a multi-level slot attention module that fuses high-level semantic features from SARATR-X and low-level scattering features from a wavelet scattering network. This fusion enables the model to distinguish structurally similar target and background regions, a core challenge in SAR imagery. By incorporating structural and spatial guidance, the proposed slot refinement process effectively suppresses spurious activations on background clutter. Extensive experiments under various extended operating conditions, including unseen scenes and sensor parameters, demonstrate that the semantic features and the structural cues can complement each other’s information for object-centric learning in SAR imagery, which is what enables the proposed model to significantly outperform current state-of-the-art methods.

\bibliographystyle{unsrt}
\bibliography{ref}

\begin{thebibliography}{10}

\bibitem{cumming}
Ian~G Cumming and Frank~H Wong.
\newblock Digital processing of synthetic aperture radar data.
\newblock {\em Artech house}, 1(3):108--110, 2005.

\bibitem{atrnet}
Yongxiang Liu, Weijie Li, Li~Liu, Jie Zhou, Bowen Peng, Yafei Song, Xuying Xiong, Wei Yang, Tianpeng Liu, Zhen Liu, et~al.
\newblock Atrnet-star: A large dataset and benchmark towards remote sensing object recognition in the wild.
\newblock {\em arXiv preprint arXiv:2501.13354}, 2025.

\bibitem{stamatis}
Odysseas Kechagias-Stamatis and Nabil Aouf.
\newblock Automatic target recognition on synthetic aperture radar imagery: A survey.
\newblock {\em IEEE Aerospace and Electronic Systems Magazine}, 36(3):56--81, 2021.

\bibitem{zhou}
Jie Zhou, Chao Xiao, Bo~Peng, Zhen Liu, Li~Liu, Yongxiang Liu, and Xiang Li.
\newblock Diffdet4sar: Diffusion-based aircraft target detection network for sar images.
\newblock {\em IEEE Geoscience and Remote Sensing Letters}, 21:1--5, 2024.

\bibitem{sardet100k}
Yuxuan Li, Xiang Li, Weijie Li, Qibin Hou, Li~Liu, Ming-Ming Cheng, and Jian Yang.
\newblock Sardet-100k: Towards open-source benchmark and toolkit for large-scale sar object detection.
\newblock {\em Advances in Neural Information Processing Systems}, 37:128430--128461, 2024.

\bibitem{saratrx}
Weijie Li, Wei Yang, Yuenan Hou, Li~Liu, Yongxiang Liu, and Xiang Li.
\newblock Saratr-x: Towards building a foundation model for sar target recognition.
\newblock {\em IEEE Transactions on Image Processing}, 2025.

\bibitem{clutterregion}
Maria~S Greco and Fulvio Gini.
\newblock Statistical analysis of high-resolution sar ground clutter data.
\newblock {\em IEEE Transactions on Geoscience and Remote sensing}, 45(3):566--575, 2007.

\bibitem{irasnet}
Oh-Tae Jang, Min-Jun Kim, Sung-Ho Kim, Hee-Sub Shin, and Kyung-Tae Kim.
\newblock Irasnet: Improved feature-level clutter reduction for domain generalized sar-atr.
\newblock {\em IEEE Transactions on Aerospace and Electronic Systems}, pages 1--18, 2025.

\bibitem{hdanet}
Weijie Li, Wei Yang, Wenpeng Zhang, Tianpeng Liu, Yongxiang Liu, and Li~Liu.
\newblock Hierarchical disentanglement-alignment network for robust sar vehicle recognition.
\newblock {\em IEEE Journal of Selected Topics in Applied Earth Observations and Remote Sensing}, 16:9661--9679, 2023.

\bibitem{cfa}
Bowen Peng, Jianyue Xie, Bo~Peng, and Li~Liu.
\newblock Learning invariant representation via contrastive feature alignment for clutter robust sar atr.
\newblock {\em IEEE Geoscience and Remote Sensing Letters}, 20:1--5, 2023.

\bibitem{tsmal}
Shuai Guo, Ting Chen, Penghui Wang, Junkun Yan, and Hongwei Liu.
\newblock Tsmal: Target-shadow mask assistance learning network for sar target recognition.
\newblock {\em IEEE Journal of Selected Topics in Applied Earth Observations and Remote Sensing}, 17:18247--18263, 2024.

\bibitem{ocl1}
Klaus Greff, Sjoerd Van~Steenkiste, and J{\"u}rgen Schmidhuber.
\newblock Neural expectation maximization.
\newblock {\em Advances in neural information processing systems}, 30, 2017.

\bibitem{sa}
Francesco Locatello, Dirk Weissenborn, Thomas Unterthiner, Aravindh Mahendran, Georg Heigold, Jakob Uszkoreit, Alexey Dosovitskiy, and Thomas Kipf.
\newblock Object-centric learning with slot attention.
\newblock {\em Advances in neural information processing systems}, 33:11525--11538, 2020.

\bibitem{dinosaur}
Maximilian Seitzer, Max Horn, Andrii Zadaianchuk, Dominik Zietlow, Tianjun Xiao, Carl-Johann Simon-Gabriel, Tong He, Zheng Zhang, Bernhard Sch{\"o}lkopf, Thomas Brox, et~al.
\newblock Bridging the gap to real-world object-centric learning.
\newblock {\em arXiv preprint arXiv:2209.14860}, 2022.

\bibitem{isa}
Ondrej Biza, Sjoerd Van~Steenkiste, Mehdi~SM Sajjadi, Gamaleldin~F Elsayed, Aravindh Mahendran, and Thomas Kipf.
\newblock Invariant slot attention: Object discovery with slot-centric reference frames.
\newblock {\em arXiv preprint arXiv:2302.04973}, 2023.

\bibitem{ocl2}
Horace Barlow.
\newblock Grandmother cells, symmetry, and invariance: how the term arose and what the facts suggest.
\newblock 2009.

\bibitem{aaaisar}
Xi~Yang, Jiachen Sun, Songsong Duan, and De~Cheng.
\newblock Dual information purification for lightweight sar object detection.
\newblock In {\em Proceedings of the AAAI Conference on Artificial Intelligence}, volume~39, pages 9274--9282, 2025.

\bibitem{cvprsar}
Shasha Mao, Shiming Lu, Zhaolong Du, Licheng Jiao, Shuiping Gou, Luntian Mou, Xuequan Lu, Lin Xiong, and Yimeng Zhang.
\newblock Cross-rejective open-set sar image registration.
\newblock In {\em Proceedings of the Computer Vision and Pattern Recognition Conference}, pages 23027--23036, 2025.

\bibitem{gsa}
Minhyeok Lee, Suhwan Cho, Dogyoon Lee, Chaewon Park, Jungho Lee, and Sangyoun Lee.
\newblock Guided slot attention for unsupervised video object segmentation.
\newblock In {\em Proceedings of the IEEE/CVF conference on computer vision and pattern recognition}, pages 3807--3816, 2024.

\bibitem{wsn}
Shanel Gauthier, Benjamin Th{\'e}rien, Laurent Alsene-Racicot, Muawiz Chaudhary, Irina Rish, Eugene Belilovsky, Michael Eickenberg, and Guy Wolf.
\newblock Parametric scattering networks.
\newblock In {\em Proceedings of the IEEE/CVF Conference on Computer Vision and Pattern Recognition}, pages 5749--5758, 2022.

\bibitem{jongsenlee}
Jong-Sen Lee.
\newblock {Speckle Suppression and Analysis for Synthetic Aperture Radar Images}.
\newblock In Henri~H. Arsenault, editor, {\em Intl Conf on Speckle}, volume 0556, pages 170 -- 179. International Society for Optics and Photonics, SPIE, 1985.

\bibitem{Frost}
Victor~S. Frost, Josephine~Abbott Stiles, K.~S. Shanmugan, and Julian~C. Holtzman.
\newblock A model for radar images and its application to adaptive digital filtering of multiplicative noise.
\newblock {\em IEEE Transactions on Pattern Analysis and Machine Intelligence}, PAMI-4(2):157--166, 1982.

\bibitem{potter1997attributed}
Lee~C Potter and Randolph~L Moses.
\newblock Attributed scattering centers for sar atr.
\newblock {\em IEEE Transactions on image processing}, 6(1):79--91, 1997.

\bibitem{zhang2023vsfa}
Chen Zhang, Yinghua Wang, Hongwei Liu, Yuanshuang Sun, and Siyuan Wang.
\newblock Vsfa: Visual and scattering topological feature fusion and alignment network for unsupervised domain adaptation in sar target recognition.
\newblock {\em IEEE Transactions on Geoscience and Remote Sensing}, 61:1--20, 2023.

\bibitem{zhao2024adaptive}
Chenxi Zhao, Daochang Wang, Xianghui Zhang, Yuli Sun, Siqian Zhang, and Gangyao Kuang.
\newblock Adaptive scattering feature awareness and fusion for limited training data sar target recognition.
\newblock {\em IEEE Journal of Selected Topics in Applied Earth Observations and Remote Sensing}, 2024.

\bibitem{busip}
Chenxi Zhao, Daochang Wang, Siqian Zhang, and Gangyao Kuang.
\newblock Bottom-up scattering information perception network for sar target recognition.
\newblock {\em arXiv preprint arXiv:2504.04780}, 2025.

\bibitem{ifts}
Jae-Ho Choi, Myung-Jun Lee, Nam-Hoon Jeong, Geon Lee, and Kyung-Tae Kim.
\newblock Fusion of target and shadow regions for improved sar atr.
\newblock {\em IEEE Transactions on Geoscience and Remote Sensing}, 60:1--17, 2022.

\bibitem{lmbncnn}
Feng Zhou, Li~Wang, Xueru Bai, and Ye~Hui.
\newblock Sar atr of ground vehicles based on lm-bn-cnn.
\newblock {\em IEEE Transactions on Geoscience and Remote Sensing}, 56(12):7282--7293, 2018.

\bibitem{smsa}
Dongwon Kim, Seoyeon Kim, and Suha Kwak.
\newblock Bootstrapping top-down information for self-modulating slot attention.
\newblock {\em Advances in Neural Information Processing Systems}, 37:103751--103773, 2024.

\bibitem{scholkopf2021toward}
Bernhard Sch{\"o}lkopf, Francesco Locatello, Stefan Bauer, Nan~Rosemary Ke, Nal Kalchbrenner, Anirudh Goyal, and Yoshua Bengio.
\newblock Toward causal representation learning.
\newblock {\em Proceedings of the IEEE}, 109(5):612--634, 2021.

\bibitem{dittadi2021generalization}
Andrea Dittadi, Samuele Papa, Michele De~Vita, Bernhard Sch{\"o}lkopf, Ole Winther, and Francesco Locatello.
\newblock Generalization and robustness implications in object-centric learning.
\newblock {\em arXiv preprint arXiv:2107.00637}, 2021.

\bibitem{jiang2023object}
Jindong Jiang, Fei Deng, Gautam Singh, and Sungjin Ahn.
\newblock Object-centric slot diffusion.
\newblock {\em arXiv preprint arXiv:2303.10834}, 2023.

\bibitem{wu2023slotdiffusion}
Ziyi Wu, Jingyu Hu, Wuyue Lu, Igor Gilitschenski, and Animesh Garg.
\newblock Slotdiffusion: Object-centric generative modeling with diffusion models.
\newblock {\em Advances in Neural Information Processing Systems}, 36:50932--50958, 2023.

\bibitem{kim2023improving}
Dongwon Kim, Namyup Kim, and Suha Kwak.
\newblock Improving cross-modal retrieval with set of diverse embeddings.
\newblock In {\em Proceedings of the IEEE/CVF conference on computer vision and pattern recognition}, pages 23422--23431, 2023.

\bibitem{jsa}
Inho Kim, Youngkil Song, Jicheol Park, Won~Hwa Kim, and Suha Kwak.
\newblock Improving sound source localization with joint slot attention on image and audio.
\newblock In {\em Proceedings of the Computer Vision and Pattern Recognition Conference}, pages 3121--3130, 2025.

\bibitem{liao2025future}
Guiqiu Liao, Matjaz Jogan, Marcel Hussing, Edward Zhang, Eric Eaton, and Daniel~A Hashimoto.
\newblock Future slot prediction for unsupervised object discovery in surgical video.
\newblock {\em arXiv preprint arXiv:2507.01882}, 2025.

\bibitem{kipf2021conditional}
Thomas Kipf, Gamaleldin~F Elsayed, Aravindh Mahendran, Austin Stone, Sara Sabour, Georg Heigold, Rico Jonschkowski, Alexey Dosovitskiy, and Klaus Greff.
\newblock Conditional object-centric learning from video.
\newblock {\em arXiv preprint arXiv:2111.12594}, 2021.

\bibitem{yu2024multimodal}
Tianyi Yu, Fangzhou Han, Lamei Zhang, and Bin Zou.
\newblock Multimodal slot vision transformer for sar image classification.
\newblock In {\em 2024 IEEE International Conference on Signal, Information and Data Processing (ICSIDP)}, pages 1--6. IEEE, 2024.

\bibitem{han2025slotfusion}
Fangzhou Han, Tianyi Yu, Lamei Zhang, Lingyu Si, and Yiqi Zhang.
\newblock Slotfusion: Object-centric audiovisual feature fusion with slot attention for remote sensing scene recognition.
\newblock In {\em ICASSP 2025-2025 IEEE International Conference on Acoustics, Speech and Signal Processing (ICASSP)}, pages 1--5. IEEE, 2025.

\bibitem{WT}
I.~Daubechies.
\newblock The wavelet transform, time-frequency localization and signal analysis.
\newblock {\em IEEE Transactions on Information Theory}, 36(5):961--1005, 1990.

\bibitem{Duan}
Yiping Duan, Fang Liu, Licheng Jiao, Peng Zhao, and Lu~Zhang.
\newblock Sar image segmentation based on convolutional-wavelet neural network and markov random field.
\newblock {\em Pattern Recognition}, 64:255--267, 2017.

\bibitem{Tello}
M.~Tello, C.~Lopez-Martinez, and J.J. Mallorqui.
\newblock A novel algorithm for ship detection in sar imagery based on the wavelet transform.
\newblock {\em IEEE Geoscience and Remote Sensing Letters}, 2(2):201--205, 2005.

\bibitem{Grandi1}
Gianfranco~D. De~Grandi, Jong-Sen Lee, and Dale~L. Schuler.
\newblock Target detection and texture segmentation in polarimetric sar images using a wavelet frame: Theoretical aspects.
\newblock {\em IEEE Transactions on Geoscience and Remote Sensing}, 45(11):3437--3453, 2007.

\bibitem{Grandi2}
Gianfranco~D. De~Grandi, Richard~M. Lucas, and Jan Kropacek.
\newblock Analysis by wavelet frames of spatial statistics in sar data for characterizing structural properties of forests.
\newblock {\em IEEE Transactions on Geoscience and Remote Sensing}, 47(2):494--507, 2009.

\bibitem{unet}
Olaf Ronneberger, Philipp Fischer, and Thomas Brox.
\newblock U-net: Convolutional networks for biomedical image segmentation.
\newblock In {\em International Conference on Medical image computing and computer-assisted intervention}, pages 234--241. Springer, 2015.

\bibitem{gru}
Junyoung Chung, Caglar Gulcehre, KyungHyun Cho, and Yoshua Bengio.
\newblock Empirical evaluation of gated recurrent neural networks on sequence modeling.
\newblock {\em arXiv preprint arXiv:1412.3555}, 2014.

\bibitem{adam}
Diederik~P Kingma and Jimmy Ba.
\newblock Adam: A method for stochastic optimization.
\newblock {\em arXiv preprint arXiv:1412.6980}, 2014.

\bibitem{oquab2023dinov2}
Maxime Oquab, Timoth{\'e}e Darcet, Th{\'e}o Moutakanni, Huy Vo, Marc Szafraniec, Vasil Khalidov, Pierre Fernandez, Daniel Haziza, Francisco Massa, Alaaeldin El-Nouby, et~al.
\newblock Dinov2: Learning robust visual features without supervision.
\newblock {\em arXiv preprint arXiv:2304.07193}, 2023.

\bibitem{tsne}
Laurens van~der Maaten and Geoffrey Hinton.
\newblock Visualizing data using t-sne.
\newblock {\em Journal of machine learning research}, 9(Nov):2579--2605, 2008.

\end{thebibliography}

\end{document}